\documentclass[10pt,twocolumn,letterpaper]{article}

\usepackage[pagenumbers]{wacv} 

\usepackage{graphicx}
\usepackage{amsmath}
\usepackage{amssymb}
\usepackage{booktabs}
\usepackage[T1]{fontenc}
\usepackage[dvipsnames]{xcolor}
\usepackage{multirow}

\newcommand{\rotationangle}{0}

\usepackage[pagebackref,breaklinks,colorlinks]{hyperref}

\usepackage[capitalize]{cleveref}
\crefname{section}{Sec.}{Secs.}
\Crefname{section}{Section}{Sections}
\Crefname{table}{Table}{Tables}
\crefname{table}{Tab.}{Tabs.}

\def\wacvPaperID{0618} 
\def\confName{WACV}
\def\confYear{2025}

\begin{document}

\title{Label Convergence: Defining an Upper Performance Bound in Object Recognition through Contradictory Annotations}

\author{
    David Tschirschwitz \quad Volker Rodehorst\\ \\
    Bauhaus-Universität Weimar, Germany\\
    {\tt\small david.tschirschwitz@uni-weimar.de} 
}
\maketitle

\begin{abstract}
    Annotation errors are a challenge not only during training of machine learning models, but also during their evaluation. Label variations and inaccuracies in datasets often manifest as contradictory examples that deviate from established labeling conventions. Such inconsistencies, when significant, prevent models from achieving optimal performance on metrics such as mean Average Precision (mAP). We introduce the notion of ``label convergence`` to describe the highest achievable performance under the constraint of contradictory test annotations, essentially defining an upper bound on model accuracy.

    Recognizing that noise is an inherent characteristic of all data, our study analyzes five real-world datasets, including the LVIS dataset, to investigate the phenomenon of label convergence. We approximate that label convergence is between 62.63-67.52 mAP@[0.5:0.95:0.05] for LVIS with 95\% confidence, attributing these bounds to the presence of real annotation errors. With current state-of-the-art (SOTA) models at the upper end of the label convergence interval for the well-studied LVIS dataset, we conclude that model capacity is sufficient to solve current object detection problems. Therefore, future efforts should focus on three key aspects: (1) updating the problem specification and adjusting evaluation practices to account for unavoidable label noise, (2) creating cleaner data, especially test data, and (3) including multi-annotated data to investigate annotation variation and make these issues visible from the outset.
\end{abstract}

\section{Introduction}
\label{sec:intro}

\begin{figure}[ht]
\begin{center}
    \includegraphics[width=\linewidth]{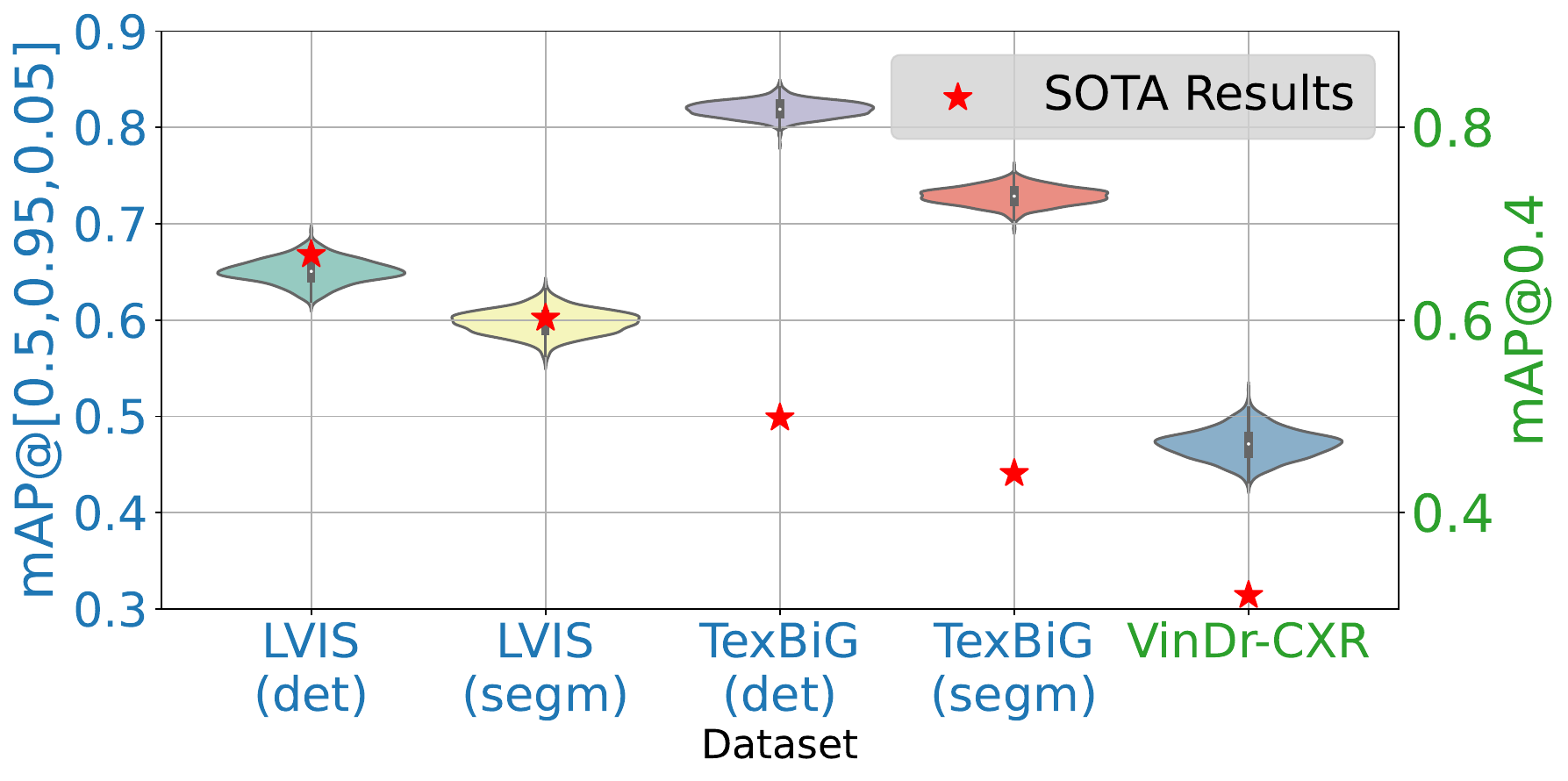}
    \caption{Illustration of the convergence threshold intervals and the respective state-of-the-art (SOTA) results, highlighting how close the data points are to the upper performance bound -- at least for LVIS. For the LVIS dataset, the convergence threshold and SOTA results (using pre-trained Co-DETR \cite{co_detr}) are evaluated on the consistency subset of LVIS. The convergence threshold interval for LVIS is created directly using modified mAP@[0.5,0.95,0.05] as described in Section \ref{sec:two-annos}. For the TexBiG and VinDr-CXR datasets, the convergence threshold interval is inferred using the formula from K-$\alpha$ to mAP introduced in Section \ref{sec:mul-annos}. The TexBiG and VinDr-CXR datasets utilize their respective leaderboard results, with TexBiG using modified mAP@[0.5,0.95,0.05] and VinDr-CXR using the mAP@0.4 metric as used in their respective leaderboards.}
    \label{fig:convergence_threshold}
\end{center}
\end{figure}

\begin{table*}[ht]
\centering
\begin{tabular}{@{}lrcrccc@{}}
\toprule
Name                  & Images & Classes & Instance & \# Rat./Img. & Masks & Guideline \\ \midrule
LVIS Cons. Subset \cite{lvis}       & 5,000  & 1203     & 100,480  & 2.00        & Yes   & Same     \\
COCO Reann. \cite{ma2022}           & 80,067 & 5       & 1,022,716& 2.00        & Yes   & Different      \\
Open Images Reann. \cite{ma2022}    & 4,773  & 5       & 31,198   & 2.00        & No    & Different      \\
TexBiG \cite{tschirschwitz2022}                & 2,257  & 19      & 53,623   & 2.16        & Yes   & Same     \\
VinDr-CXR \cite{nguyen2022a}            & 15,000 & 14      & 36,557   & 3.00        & No    & Same     \\ \bottomrule
\end{tabular}
\caption{Datasets with repeated labels, that enable the analysis of annotation variation. TexBiG and VinDr-CXR have repeated labels for the full dataset. For LVIS, only 5,000 and/or the 19,809 validation data are used because they were annotated by multiple persons for consistency measurement. COCO and Open Images use the original and reannotated versions combined to identify label variation, but use different guidelines and have some inconsistencies in the organization of the annotations as described in \hyperref[sec:appendix-A]{Appendix A}.}
\label{tab:datasets-summary}
\end{table*}

Machine learning systems can be categorized into three broad phases: (1) dataset creation, (2) model development, and (3) evaluation. Within the computer vision community, there is a significant focus on the second phase—the development of innovative methods to address new challenges or improve results for existing problems. In such cases, the training data, test data, and evaluation criteria remain constant, and the sole focus is on model modifications and adaptations to enhance system performance. In contrast to this model-centric paradigm, there exists an alternative approach known as data-centric AI. This concept was introduced by Andrew Ng \cite{deeplearningai2021} and primarily concentrates on the first phase. This strategy involves maintaining a constant model while systematically enhancing the quality of the data. Within this data-centric approach, improvements focus on aspects such as label consistency, the guidelines or conventions followed by annotators\footnote{We consider the guideline as the instructions given to the annotators and the annotation convention, the interpretation of annotators made, which in the best case exactly matches the guideline if the guideline is unambiguous.}, enhancing domain coverage, adjusting the dataset size, and identifying errors, among others. Both methods aim to boost performance, either through advancements in data quality or through model optimization.

\par
In this study, we adopt an alternate perspective by including an additional ``zeroth`` phase -- problem specification (0). Our focus is on both the problem specification (0) and the evaluation (3), and we frame our investigation around the following central question:

\begin{center}
\textit{"How can we estimate the intrinsic performance threshold for models and data given annotation variation in human labels such as noise or uncertainty?"}
\end{center}

This study introduces the concept of \textit{label convergence}, which hypothesizes that the performance limitations of a dataset arise from internal label inconsistencies. These inconsistencies may stem from variations in labeling conventions, annotator variability, or outright errors. Many of these challenges originate in the problem specification phase, where datasets typically assume the provided labels are ``gold standard``, representing a singular ground truth. However, for complex data with unavoidable label noise, perfect ground truth is often unattainable. In such cases, label convergence serves as a measure of the inherent ambiguity in both problem specification and evaluation. Before delving into this topic, we first review the significance of labels in machine learning.

\par
The issues with annotations are referred to by several terms: noisy labels, human label variation, annotation errors, or uncertain labels. We will refer to them as (annotation/label) \textit{variations}. Despite the different terminologies, they all denote the same underlying concept where the mapping from a feature vector to a label is not uniquely defined. This phenomenon can be expressed for a set of images, denoted by $\{x_i\}^N_{i=1}$, where an annotator has produced a set of labels $\{\Tilde{y}_{ij}\}^{M}_{j=0}$ for each image $i$. Here, $j$ refers to individual instances within an image, such as a bounding box or a segmentation mask, that are intended to represent specific objects or segments. These annotated labels aim to approximate the true but unknown labels $y_{ij}$ \cite{tanno2019,tschirschwitz2023a}.

The detrimental effects of label variation on the training process have been extensively studied, with results indicating significant performance degradation \cite{bar2023}, negative effects on the training process \cite{mao2021a}, and limitations on the accuracy of learning algorithms \cite{tanno2019}. It has been highlighted that among various types of noise, label variation is particularly damaging. This assertion is supported by evidence of a substantial performance difference between training on clean versus varying data, as demonstrated in the research by Song \etal \cite{song2022}. Additionally, a study by Xu \etal \cite{xu2019} revealed that missing 20\% of the instances for training can lead to a drop in performance of about 5 percentage points (from 70 to 65 mAP on PASCAL VOC 2007 \cite{pascal_voc}). This form of noise is of particular concern because deep neural networks have a significant memory capacity, which can lead to overfitting to varying labels if they are prevalent in the training dataset \cite{zhang2020a}. To reduce the impact of varying labels, it may be preferable to use smaller capacity models but train on cleaner labels \cite{northcutt2021,song2022}, especially since performance improvements are only logarithmic with increasing dataset size \cite{sun2017}.

\par

According to research referenced by Song \etal \cite{song2022}, the prevalence of noisy or incorrect labels in "real-world datasets" is estimated to range from 8.0\% to 38.0\%. These results primarily concern the domain of image classification. However, to the best of our knowledge, and as supported by Schubert \etal \cite{schubert2024} and Agnew \etal \cite{agnew2023}, the extent of label variation within object detection datasets remains undetermined.

\par

Training with varying labels has been categorized into four distinct strategies by Plank \cite{plank2022}, who specifically addresses annotation variation as human label variation, with a notable emphasis on natural language processing (NLP). These strategies are divided into two main objectives: (1) addressing human label variation through either (1.1) aggregating labels or (1.2) filtering out noise, and (2) leveraging human label variation by either (2.1) learning directly from unaggregated labels or (2.2) integrating gold standard labels with human label variation.

\begin{figure*}[ht]
\begin{center}
    \includegraphics[width=0.95\textwidth]{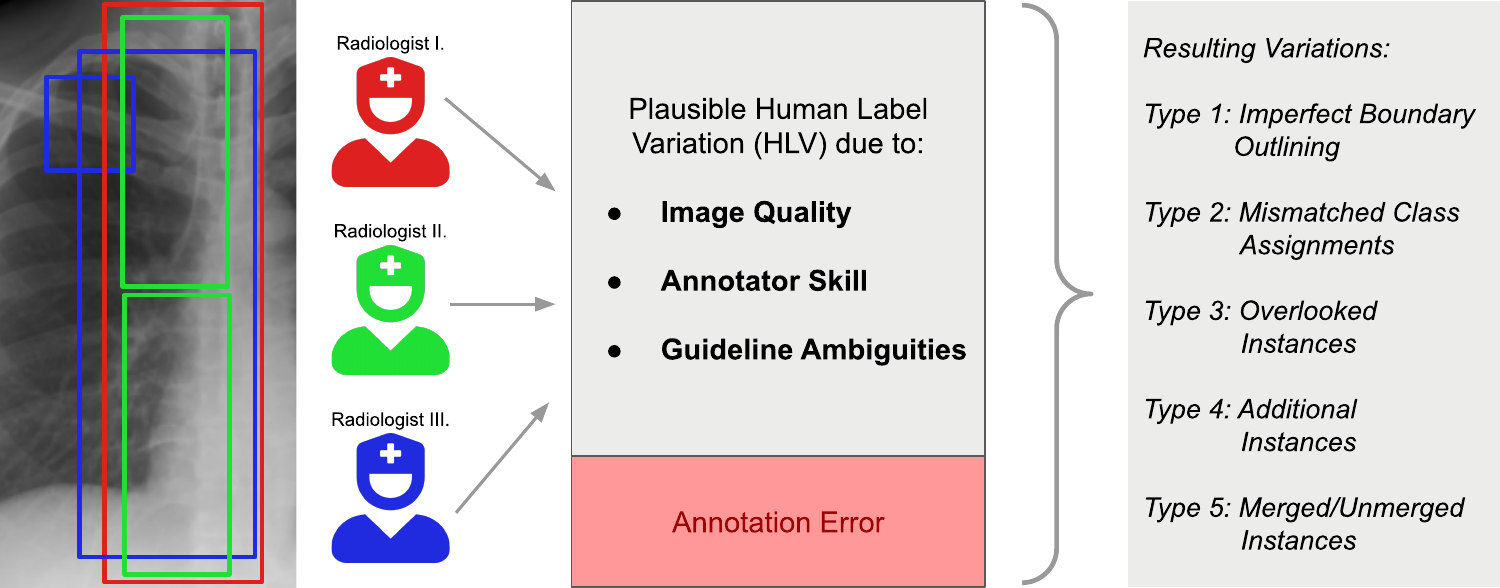}
    \caption{Human label variation causes and annotation errors causing five different types of variations. Our graphic is inspired by Plank \cite{plank2022}. The fifth error type is illustrated between the green and red radiologist.}
    \label{fig:label_variation_types}
\end{center}
\end{figure*}

The issue of learning with varying annotations has received attention in the research community, but often adopts a model-centric AI perspective, especially with respect to the test data. This approach typically assumes that the test data are either of sufficient quality, or that any flaws within the test data are deliberately overlooked in order to prioritize model improvements. In this realm, strategies for managing annotation variation often involve the synthetic introduction of variation \cite{li2020d,bar2023,liu2022a}, aiming for mitigation through techniques such as co-teaching \cite{chadwick2019}, soft-label learning \cite{li2020c,hu2019,peng2020}, or noise filtering \cite{mao2021,mao2021a}. 

A subset of methods also tackles the real-world label variation that occurs in datasets with ``repeated labels`` -- that is, labels provided by multiple annotators for the same image. Extending the earlier mathematical formulation, such repeated annotations would be denoted by $\Tilde{y}_{ij}^{r}$, where $r = 1, \ldots, R$ represents the index of the annotator. In the context of object detection and instance segmentation, two significant datasets have been introduced: the VinDr-CXR dataset \cite{nguyen2022a} and the TexBiG dataset \cite{tschirschwitz2022}. Research efforts focused that have shown results on these datasets with real label variations \cite{tschirschwitz2023a,pham2023,pham2022,h.le2023} primarily explore label aggregation techniques to approximate the true label. 

\par

While the focus of many studies on varying labels has been on enhancing the training process, it is the test data that ultimately determine the system's utility in practice. For a system to be viable, it's imperative that the test data comprehensively represent the problem domain, thereby minimizing the discrepancy between controlled (in-vitro) and real-world (in-vivo) conditions \cite{plank2022}. This discussion circles back to the concept of \textit{label convergence} introduced earlier, emphasizing the critical nature of precise and consistent labeling and if that is not further possible a problem specification that accounts for the remaining uncertainties.

When considering the application of neural networks, we distinguish between three types of upper bounds that can affect performance:

\begin{enumerate}
    \item \textbf{Model Convergence}: This refers to the ability of a model to learn from data and generalize to unseen data. Key issues include model bias and variance, which are affected by factors such as optimization techniques, network architecture, and other parameters.
    \item \textbf{Data Convergence}: This involves the completeness of the dataset. Models typically struggle to learn from data outside their training domain, which affects their ability to generalize. This category also includes considerations of data augmentation techniques.
    \item \textbf{Label Convergence}: This hypothesis suggests that dataset performance is inherently limited by conflicting labels due to annotation variation. These inconsistencies can significantly hinder the learning process.
\end{enumerate}

In addition to these types of convergence, system performance may also be constrained by hardware limitations or other external factors. Our contributions to the understanding of label convergence in object recognition include:
\begin{itemize}
    \item We present a method for evaluating the convergence threshold interval using a modified mAP, allowing direct comparison with model results and determining the upper-performance bound. This method is applied to estimate the convergence threshold for the LVIS dataset.
    \item We extend this method to multiple annotators by correlating the modified mAP with an inter-annotator agreement metric and estimate the convergence threshold for the VinDr-CXR and TexBiG datasets.
    \item We provide the first analysis of real annotation variation within object detection and instance segmentation, and offer a tangible distribution of this annotation variation across five datasets.
\end{itemize}

\section{Related Work}
\label{sec:rel_work}

Previous studies have approached the concept of label convergence indirectly, though not explicitly named as such. These efforts, aimed at understanding the upper bounds of model performance, have generally relied on the analysis of learned models. However, assessing label convergence ideally requires a model-independent approach that focuses solely on label quality.

Borji \etal \cite{borji2019} investigated an "Empirical Upper Bound" for object detection models, potentially touching on aspects of the three types of convergence we've discussed. They hypothesized an upper bound in mAP determined by the performance of an ideal detector using ground truth bounding boxes and the optimal object classifier. While intriguing, their methodology overlooks localization and relies on learned models, which deviates from a pure label convergence evaluation, which should be independent of any model.

Agnew \etal \cite{agnew2023} attempted to quantify annotation quality empirically by introducing synthetic variations into the annotations. Their methods include uniform noise for expanding bounding boxes and Gaussian radial noise for polygons. While the objective is to identify a performance ceiling due to contradictory examples, the reliance on synthetic noise and learned models limits the study. These noise patterns, being predictable, are significantly easier to learn by models \cite{merdjanovska2024}, and the assessment is not fully detached from the original, potentially flawed, annotations.

Ma \etal \cite{ma2022} adopted a data-centric AI approach by re-annotating datasets such as MS COCO and Google Open Images. By comparing model performance across various combinations of old and new annotations, they explored the impact of different labeling conventions. While their work points to the interplay between model, data, and label convergence, it again depends on trained models, making it a partial attempt at empirical label convergence evaluation.

In summary, while these studies illuminate challenges of annotation, they fall short of defining label convergence because they rely on learned models and their primary objectives do not align with a pure assessment of label quality.

\par

A promising approach for assessing label convergence is to examine annotation consistency, similar to how performance metrics evaluate model quality. While measuring annotation consistency is common in natural language processing data collections \cite{plank2022}, major datasets such as MS COCO \cite{coco} lack such consistency metrics. Conversely, datasets like LVIS \cite{lvis} attempt to measure consistency but use techniques designed to evaluate model predictions against ground truth, such as an F1-score. Even the most common metrics like mAP used for COCO would not work out of the box for several reasons: 1) the ranking of detections in model evaluation relies on confidence scores, which are absent in repeated labeling scenarios because no prediction score is provided, and 2) precision and recall measures are impractical without a definitive count of total detections. Typically, consistency is assessed using statistical measures of inter-annotator agreement, such as Cohen's Kappa \cite{cohen1960coefficient} or Krippendorff's Alpha \cite{krippendorff_computing, krippendorff_pros}. However, adapting these methods for object detection or instance segmentation is challenging due to the need to incorporate localization.

Existing methods for semantic segmentation, such as Ribeiro \etal \cite{ribeiro2019}, apply Cohen's Kappa on a per-pixel basis, but overlook the distinction between instances. Similarly, Nassar \etal \cite{nassar2019} conduct a pixel-wise inter-annotator evaluation using Krippendorff's Alpha for object detection, again neglecting instance-specific considerations. To our knowledge, the only peer-reviewed approach to evaluating annotation consistency for object detection that considers the instance-based nature has been developed for the TexBiG \cite{tschirschwitz2022} dataset. It uses a unique method that combines Intersection over Union (IoU) thresholding with graph-based matching to address localization before applying a conventional Krippendorff's Alpha for assessment.

\par

In addition to assessing the consistency of annotations, several studies have analyzed the types of variations prevalent in object detection or instance segmentation annotations, which ultimately result in the label convergence threshold. A detailed classification of these variations is illustrated in Figure \ref{fig:label_variation_types}, which includes five main categories:

\begin{enumerate}
    \item \textbf{Imperfect Boundary Outlining (Bad Boundary):} This variation reflects a mismatch between the actual bounding box or mask and its annotated counterpart. Minor deviations are almost inevitable in all annotations and can be considered negligible.
    \item \textbf{Mismatched Class Assignment (Wrong Class):} This occurs when the annotation assigns an incorrect class to an object, despite accurate localization.
    \item \textbf{Overlooked Instances (Missed Instances):} Instances that are present in the image but not in the annotation are overlooked instances. In model evaluation, this would be similar to a false negative.
    \item \textbf{Additional Instance:} Annotated instances without a corresponding object in the image, or for classes not represented in the dataset. This error is equivalent to a false positive in the model evaluation.
    \item \textbf{Merged or Unmerged Instances:} Discrepancies in annotator decisions about whether to merge or separate instances, often influenced by occlusion or other factors, resulting in unnecessary merging or splitting of instances.
\end{enumerate}

Our study contrasts with three other studies that simulate label variation. In Table \ref{tab:error-analysis}, we compare the variations analyzed in both our study and theirs, demonstrating the comprehensive nature of our approach in examining real dataset variations across repeated labels in contrast to other studies that rely on synthetic label variations.

\begin{table}[ht]
\footnotesize
\centering
\begin{tabular}{@{}lcccc@{}}
\toprule
 &
\rotatebox{\rotationangle}{\begin{tabular}[c]{@{}c@{}}Chan \\ \etal \cite{chan2023}\end{tabular}} &
\rotatebox{\rotationangle}{\begin{tabular}[c]{@{}c@{}}Schubert \\ \etal \cite{schubert2024}\end{tabular}} &
\rotatebox{\rotationangle}{\begin{tabular}[c]{@{}c@{}}Chachula \\ \etal \cite{chachula2023}\end{tabular}} &
\rotatebox{\rotationangle}{\begin{tabular}[c]{@{}c@{}}Ours\end{tabular}} \\
\midrule
Bad Boundaries (Type 1)   & \checkmark & \checkmark & \checkmark & \checkmark \\
Wrong Class (Type 2)      & \checkmark & \checkmark & \checkmark & \checkmark \\
Missed Instances (Type 3) & \checkmark & \checkmark & \checkmark & \checkmark \\
Additional Inst. (Type 4) & $\times$ & \checkmark & \checkmark & $\times$ \\
Merging Issue (Type 5)    & $\times$   & $\times$   & $\times$   & \checkmark \\
Redundancy (no error)     & \checkmark & $\times$   & $\times$   & $\times$ \\
\bottomrule
\end{tabular}
\caption{Comparative analysis of different approaches on handling specific annotation variation types.}
\label{tab:error-analysis}
\end{table}

Chan \etal \cite{chan2023} explore the impact of annotation variation on model performance and identify five types of variation, four of which overlap with our classification in Figure \ref{fig:label_variation_types}. They introduce specific variations such as mislabeled class/superclass and redundant annotations—the latter of which is not considered in our variation taxonomy, assuming it can be corrected by simple post-processing such as non-maxima suppression.

Chachula \etal \cite{chachula2023} propose an algorithm for assessing and improving the quality of labels in object detection datasets. Their method, which is capable of flagging and re-annotating suspicious examples, covers all the variation types depicted in our analysis (Figure \ref{fig:label_variation_types}). This approach could potentially complement our methodology to more accurately identify annotation variations.

Schubert \etal \cite{schubert2024} present an innovative technique for detecting label variation by simulating four types of variation outlined in our study and shown in Figure \ref{fig:label_variation_types}. Their method shows promise in identifying real label variation, suggesting the potential for integration with our findings to thoroughly address and correct variation in existing datasets. This approach could also potentially be added to our own methodology for identifying annotation variation. 

However, the relevance of these studies to measuring convergence thresholds for real datasets with real label variation using simulated noise patterns remains uncertain, and the effectiveness of their variation detection methods would require extensive manual verification.

\section{Analyzed Dataset}
\label{sec:datasets}

In this section, we briefly introduce the datasets used in our study, which are summarized in Table \ref{tab:datasets-summary}. A crucial requirement for these datasets is that each image must be annotated by at least two independent raters. It is not mandatory for the raters to mark instances; if they conclude that no instance of the defined class set for the dataset is present in an image, it is acceptable to leave it unannotated. Based on these strict criteria, we selected segments from three well-known datasets in object detection: LVIS \cite{lvis}, COCO \cite{coco}, and Open Images \cite{OpenImages}.

For LVIS, we use the doubly annotated subset of 5,000 images that was originally compiled to assess annotation consistency for v0.5 of the dataset. To ensure comparability with current SOTA models, we adjusted the data to align with v1.0 of the dataset, reducing the number of categories from 1723 to 1203 and excluding annotations of the removed categories. This subset serves as our primary benchmark due to its high quality, community recognition, consistent annotation guidelines, and complexity due to its wide class variety and long-tailed distribution. Unfortunately, this dataset lacks rater identification information, which divides the dataset into two generic subgroups.

Additionally, we examined two reannotated datasets introduced by Ma \etal \cite{ma2022}, combined with the original COCO and Open Images datasets, to assess label convergence. However, these datasets have limitations that affect their validity and generalizability, detailed in \hyperref[sec:appendix-A]{Appendix A}, which prevent us from determining label convergence for these datasets.

Lastly, we consider two smaller domain-specific datasets, VinDR-CXR \cite{nguyen2022a} and TexBiG \cite{tschirschwitz2022}, which have the advantage of multiple raters per image and consistent labeling conventions throughout the annotation process, including rater identification metadata. While their smaller size and specific focus limit their impact compared to the results of the LVIS dataset, they technically provide the most suitable data for our analysis.


\section{Determining Label Convergence}
\label{sec:label-convergence}

Determining label convergence is divided into two parts: (1) for cases where each image is annotated by exactly two independent raters, and (2) for cases where images are annotated by any number of raters. A key aspect in defining label convergence is to make the convergence threshold comparable to standard evaluation metrics, allowing for direct comparison with current model performance. To achieve this, we align label convergence with the mAP used in datasets such as COCO, LVIS, and PASCAL VOC. To do this, we modify the regular mAP to work with different ground truth approximations instead of model predictions and a single ground truth. Since this still only allows us to evaluate between two annotators, we use Krippendorff’s Alpha version for object detection \cite{tschirschwitz2022}, which we also extend to work on instance segmentation, to assess the consistency of multiple annotators. By correlating these two values, we can easily but effectively determine the convergence for any number of annotators.

\subsection{Two Annotators per Image}
\label{sec:two-annos}

\begin{table*}[ht]
\centering
\begin{tabular}{llrrccccccc}
\toprule
Dataset            & Task    & Img.  & Samp. size & Mean  & Std  & Min   & Max   & CI L  & CI U & SOTA \\ \midrule
\multirow{2}{*}{LVIS Cons. Subset \cite{lvis}}  & det    & \multirow{2}{*}{5,000}  & \multirow{2}{*}{500}        & 65.08 & 1.25 & 61.49 & 69.32 & 62.64 & 67.52 & 66.8\\
                                 & segm   &         &            & 59.78 & 1.26 & 55.51 & 63.79 & 57.32 & 62.24 & 60.2 \\
{\color{gray}COCO Reann. }\cite{ma2022}        & det    & 80,067 & 8,007       & 28.39 & 1.55 & 25.42 & 31.13 & 25.35 & 31.43 & / \\
{\color{gray}Open Images Reann.} \cite{ma2022} & det    & 4,773  & 477        & 20.55 & 1.64 & 15.75 & 27.35 & 17.33 & 23.75 & /\\
\multirow{2}{*}{TexBiG \cite{tschirschwitz2022}} & det    & \multirow{2}{*}{2,257}  & \multirow{2}{*}{226}        & 81.89 & 0.82 & 78.17  & 84.57 & 80.28 & 83.50 & 49.84\\
                                & segm   &         &            & 72.87 & 0.92 & 69.40  & 75.95 & 71.07 & 74.66 & 44.06\\
VinDr-CXR  \cite{nguyen2022a}   & det    & 15,000 & 1,500       & 47.08 & 1.51 & 42.27 & 52.80 & 44.12 & 50.05 & 31.4\\ \bottomrule
\end{tabular}
\caption{Statistics from bootstrapping with 1,000 samples on the five datasets analyzed, using a sampling size of 10\% of the total dataset. This includes the lower and upper 95\% confidence intervals, which we consider to be the convergence threshold interval. For LVIS, TexBiG and VinDr-CXR we also provide the latest SOTA results.}
\label{tab:summary_statistics}
\end{table*}

To adapt the mAP metric for evaluating label convergence between two annotators, we make specific modifications. Normally, mAP requires model confidence scores and a single ground truth, but for human annotations, we address these differences as follows:

\begin{itemize}
    \item We set the confidence score of all annotations to 0.99, as human annotations are unlikely to orange produce severely overlapping detections of the same class. Unlike models, which then need to remove these excess detections through post-processing methods like non-maximum suppression.
    \item We randomly switch which annotator is considered ground truth and which is considered prediction, repeating the evaluation several times to account for variability.
\end{itemize}

Despite these adjustments, the evaluation remains comparable to the original mAP method. Here are the steps to compute the modified mAP:

\begin{enumerate}
    \item Detections are sorted from highest to lowest confidence score.
    \item Each detection is matched to a ground truth instance by IoU overlap, marking it as positive if matched or negative if not.
    \item Positives and negatives are added to the Precision-Recall (PR) curve.
    \item The area under the PR curve is calculated to obtain the AP for a category and a specific IoU threshold.
    \item The AP values are averaged across all categories to obtain the mAP@[IoU threshold].
\end{enumerate}

To statistically estimate the convergence threshold, we use bootstrapping, sampling 1,000 subsets, each with 10\% of the available images of the respective dataset and evaluate the modified mAP on each of these subsets. By evaluating the 95\% confidence interval of the sampling distribution, we capture the variability of the data and infer the convergence threshold for the entire dataset. The results show a convergence threshold interval between 62.64 and 67.52 mAP for the LVIS consistency subset, with Co-DETR's performance near the upper end at 66.8 mAP (see Table \ref{tab:summary_statistics} and Figure \ref{fig:convergence_threshold}). Due to issues with the COCO Reannotated and Open Images Reannotated datasets, these results are not included in the determination of the convergence threshold (see \hyperref[sec:appendix-A]{Appendix A}).

\paragraph{Considerations for Extrapolation:}
\begin{itemize}
    \item Label convergence was estimated for a subset of the dataset, as only some images contain repeated labels. Although similar variations are expected in the remaining data, this is an extrapolation.
    \item Domain specificity of different subsets may create a domain gap between training and testing data, potentially affecting the generalizability of the convergence threshold.
    \item Labeling conventions need to be aligned. If a dominant group of annotators follows a particular labeling convention, models may overfit to that convention. Ensuring that annotators have roughly equal contributions is crucial for accurate label convergence estimation (discussed further in Section \ref{sec:variation-analysis}). It is reasonable to assume that if such a dominant labeling convention exists within the dataset, then following that labeling convention would allow a model to exceed the convergence threshold.
\end{itemize}

\subsection{Multiple Annotators per Image}
\label{sec:mul-annos}

Since modified mAP cannot be used for images annotated by three or more annotators, we use Krippendorff's Alpha for Object Detection \cite{tschirschwitz2022} to assess annotation quality. This method, parameterized with different IoU thresholds such as mAP, provides a K-$\alpha$ value representing reliability between -1 and +1. A detailed explanation of this method is available in \hyperref[sec:appendix-B]{Appendix B}.

\begin{figure}[ht]
\begin{center}
    \includegraphics[width=\linewidth]{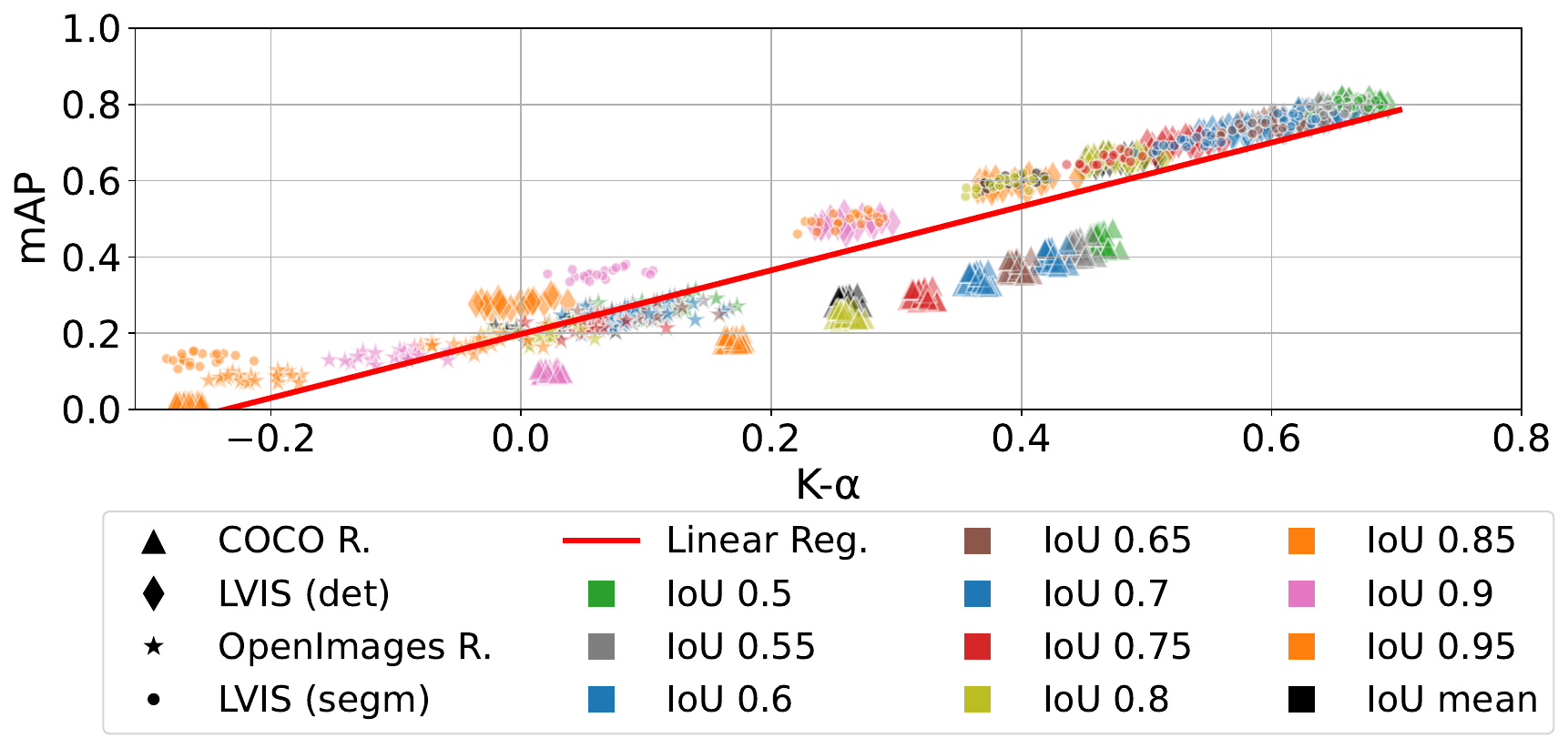}
    \caption{Best viewed digitally. Linear least-squares regression fit with K-$\alpha$ as the independent and mAP as the dependent variable. The scatter plots only show a fraction of the data points.}
    \label{fig:regression}
\end{center}
\end{figure}

To determine the convergence threshold for multiple annotators, we follow these steps:

\begin{enumerate}
    \item Use the samples from the previous bootstrapping (LVIS, COCO, Open Images) and obtain the K-$\alpha$ values for different thresholds [0.5, 0.95, 0.05] and their mean. Also evaluate the mAP for all missing IoU thresholds as well.
    \item Perform a linear least-squares regression (Figure \ref{fig:regression}) with K-$\alpha$ as the independent variable and mAP as the dependent variable. The regression shows a Pearson correlation of $\rho=0.92$ and an R-squared value $R^2=0.85$, indicating a good model fit. The resulting equation is:
    \begin{equation}
        mAP = 0.836 \cdot \alpha + 0.197
    \end{equation}
    \item Apply bootstrapping to the TexBiG and VinDr-CXR datasets to obtain K-$\alpha$@[0.5, 0.95, 0.05] for TexBiG and K-$\alpha$@[0.4] for VinDr-CXR, since these IoU thresholds were used in their respective leaderboards.
    \item Use the regression formula to infer the mAP values from the K-$\alpha$ values. Calculate the mean, standard deviation, and 95\% confidence interval to estimate the convergence threshold interval, as shown in Table \ref{tab:datasets-summary}.
\end{enumerate}


This method allows us to determine the convergence threshold for any number of annotators per image. However, these thresholds may be less reliable than directly evaluating mAP. While we use a simple linear model to describe the relationship between mAP and K-$\alpha$, future work with more data could explore better fitting models that do not assume linearity. We calculate the confidence interval empirically on the derived mAP values, as this better reflects the variability in the data than the error propagation from the regression model.

Although COCO Reannotated and Open Images Reannotated have issues (see \hyperref[sec:appendix-A]{Appendix A}), they still represent real label variations. Thus, we use these datasets to express the relationship between mAP and K-$\alpha$ in our regression calculations, preferring real variations to synthetic ones.

\section{Annotation Variation Type Analysis}
\label{sec:variation-analysis}

\begin{figure}[ht]
\begin{center}
    \includegraphics[width=\linewidth]{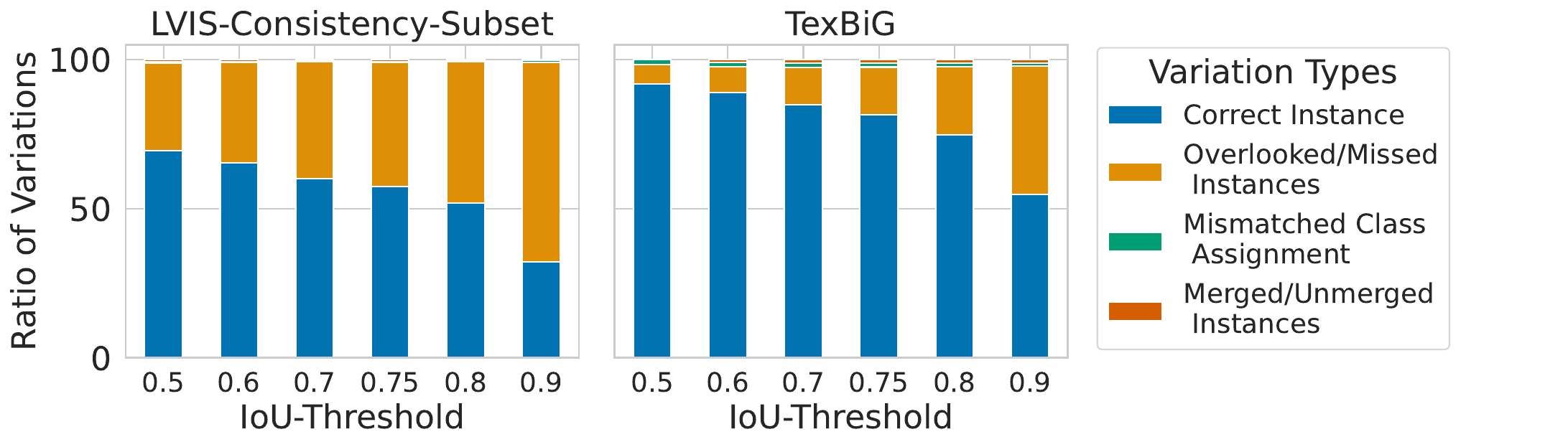}
    \caption{Quantitative results of the variation analysis on object detection, showing the ratio of correctly matched instances compared to unmatched instances. The most common variation types are 2 and 3 (Figure \ref{fig:label_variation_types}), which increase significantly as instances become harder to match due to higher IoU thresholds.}
    \label{fig:error-distribution}
\end{center}
\end{figure}

\begin{figure}[h]
    \centering
    \includegraphics[width=\linewidth]{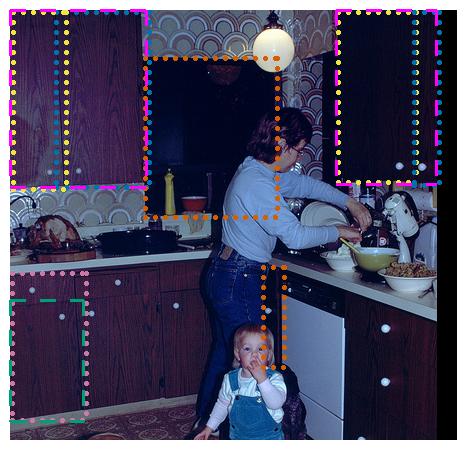}
    \caption{Example image from the LVIS Consistency Subset showing some of the variations. Only a selection of annotations from the cupboard class is visualized. The dotted line indicates coder A, while the dashed line indicates coder B. (a) At the top left and top right are two cases of a merging issue visible. This kind of inconsistencies should be prevented by a unambiguous guideline. While coder A separated the yellow and green areas of the two cupboards at the cupboard doors, coder B combined them as indicated by the magenta bounding boxes. (b) The two orange areas indicate an additional instance at the top, where a window is interpreted as a cupboard, while the bottom instance was found by coder A but missed by coder B. We attribute the first variation to image quality and the second one to an annotation error. (c) At the bottom left, another case of different labeling conventions is visible, where coder A covers the entire height of the cupboard, including the drawer, while coder B excludes the drawer from the cupboard. This could be due to either ambiguities in the guidelines or the skill of the annotator.}
    \label{fig:variation_vis}
\end{figure}

After establishing the convergence threshold, we further analyze the distribution of variation types (Figure \ref{fig:error-distribution}) and provide qualitative examples (Figure \ref{fig:variation_vis}) of real label variations. To facilitate this, we use the FiftyOne \cite{moore2020fiftyone} framework to visualize variation across the analyzed datasets. We introduce an algorithm that identifies different types of label variations and is designed to match as many examples as possible, although it tends to be strict in evaluating cases with three or more annotators. Detailed algorithmic descriptions, more qualitative examples, and an extended quantitative results section on the other datasets and instance segmentation can be found in \hyperref[sec:appendix-C]{Appendix C}.

Figure \ref{fig:variation_vis} presents a representative example of the predominant issues, illustrating the four causes of human label variation and four of the five types of variation as shown in Figure \ref{fig:label_variation_types}. This brings us back to the issue of label convergence as discussed in Section \ref{sec:two-annos}. With this tangible example, we can see what happens when annotators follow different labeling conventions. If this happens on a large scale, where multiple annotators follow a similar labeling convention that is ambiguously described in the guideline, and this group has annotated a majority of the images, models may overfit to this specific convention, potentially exceeding the presumed convergence threshold. This can be avoided by providing clear guidelines and ensuring that annotators have roughly equal contributions so that no single labeling convention dominates. Many other factors contribute to annotation variation, such as the selection of the annotation tool, which can affect perceived image quality.

\section{Conclusion}
\label{sec:conclusion}

In our study, we address our central research question by enhancing the understanding of how annotation quality impacts model performance, introducing a straightforward and effective method for determining label convergence, which establishes a theoretical upper bound on model performance.

Our analysis shows that while state-of-the-art (SOTA) results approach this upper bound for the well-studied LVIS dataset, the primary constraint is not model or data convergence but label quality. This suggests that models have sufficient capacity to handle the complexity of current object recognition problems. However, the current reliance on ``gold standard'' labels, despite inherent annotation variations, requires improvement in how problems are specified. We propose a combination of three key aspects to address this issue:

\begin{enumerate}
    \item \textbf{Improving Annotation Quality}: Implement better guidelines and training for annotators to reduce variability and errors. The issue of label convergence should be addressed as early as possible, shifting the burden from the annotations themselves to clearer problem specification. This is particularly crucial for test data, where consistency is key to accurate evaluation.

    \item \textbf{Including Multi-Annotated Data}: Since labeled data will likely always include some degree of variation, having a portion of the data annotated multiple times allows for an analysis of the extent of these variations. This enables models to become more robust to real-world scenarios by recognizing and accounting for annotation inconsistencies \cite{anand2024}.

    \item \textbf{Updating Evaluation Methods}: Reevaluate how strictly we distinguish between correct and incorrect annotations. Rather than aiming to eliminate all test set noise, we propose using the concept of label convergence as a measure of ambiguity for unavoidable annotation inconsistencies. This approach recognizes that some level of variability in labels is inevitable and should be incorporated into the evaluation process, leading to a more flexible and realistic assessment of model performance.

\end{enumerate}

In summary, our study emphasizes the critical role of label quality in achieving optimal model performance, and we propose a more nuanced approach to problem specification and evaluation that takes into account unavoidable annotation variability.

Code for our FiftyOne plugin \cite{moore2020fiftyone} can be accessed at \url{https://github.com/Madave94/multi-annotator-toolkit}.

\section{Outlook}

Our study is limited by focusing solely on computer vision, without comparisons to fields like natural language processing (NLP). We believe these domains differ significantly, as language inherently involves ambiguities, while computer vision often aims for a single, but sometimes unattainable, ground truth.

{\small
\bibliographystyle{ieee_fullname}
\bibliography{main}
}

\newpage

\section*{Appendix A: Dataset Exclusion Criteria}
\label{sec:appendix-A}

While the two reannotated datasets introduced by Ma \etal \cite{ma2022} initially provided a valuable resource for determining convergence thresholds, we encountered several issues that prevented accurate threshold determination:

\begin{enumerate}
    \item \textbf{Different Annotation Guidelines:} The datasets did not adhere to the same guidelines. Since they were annotated by different groups with varying annotation pipelines and guidelines, the annotation variations cannot be attributed to regular issues shown in Figure \ref{fig:label_variation_types}. These are not ambiguities within a single guideline but rather differences between distinct guidelines, resulting in label conventions that deviate significantly from the guideline.
    \item \textbf{Sampling Bias:} The reannotation process exhibits a sampling bias. Images were selected for reannotation based on the presence of at least one of the five chosen classes. This selection process focused on false positives while potentially overlooking false negatives, thereby skewing the dataset.
    \item \textbf{Annotation Inconsistencies:} There were inconsistencies in annotation formatting, with some annotations being untraceable to their corresponding images and vice versa. This suggests that some annotation files were incomplete.
    \item \textbf{Suspicious IoU Matches:} Anomalously high instances of perfect IoU (Intersection over Union) matches (1.0) were noted, indicating possible annotation duplication from the original datasets, although this was not explicitly confirmed in their documentation. LVIS, TexBiG, and VinDr-CXR did not contain a single instance with a 1.0 IoU overlap.
    \item \textbf{Limited Class Coverage:} Only five classes were selected for reannotation, reducing the Open Images dataset to approximately 5,000 images due to resource constraints. Extrapolating the convergence threshold from these five classes to the entire dataset decreases the validity of the estimated convergence threshold.
\end{enumerate}

Due to these points, the reannotated datasets present limited validity and generalizability. Consequently, we decided not to determine label convergence using these reannotated versions, as we do not see results on these datasets as reflective of the remaining commonly used COCO dataset. However, we still use the data to fit the linear regression, as they reflect real annotation variations, which we prefer over synthetic data.

\section*{Appendix B: Recap of Krippendorff's Alpha for Object Detection}
\label{sec:appendix-B}

To evaluate annotation consistency, we use the method introduced by Tschirschwitz \etal \cite{tschirschwitz2023a}, which adapts Krippendorff's Alpha (K-$\alpha$) for object detection. This method calculates a single $\alpha$ value to measure inter-annotator agreement, where $\alpha=1$ indicates perfect agreement, $\alpha=0$ indicates no agreement, and $\alpha<0$ indicates disagreement. The general form of K-$\alpha$ is $\alpha = 1 - \frac{D_o}{D_e}$, where $D_o$ is the observed disagreement and $D_e$ is the expected disagreement.

\subsection*{Calculation Procedure}
Using our prior definition of annotations from Section \ref{sec:intro} where a single annotation is described as $\Tilde{y}_{ij}^{r}$ which refers to annotation $j$ for image $i$ annotated by annotator $r$, the following steps are executed for a single image $i$:

\begin{figure*}[ht]
\begin{center}
    \includegraphics[width=\linewidth]{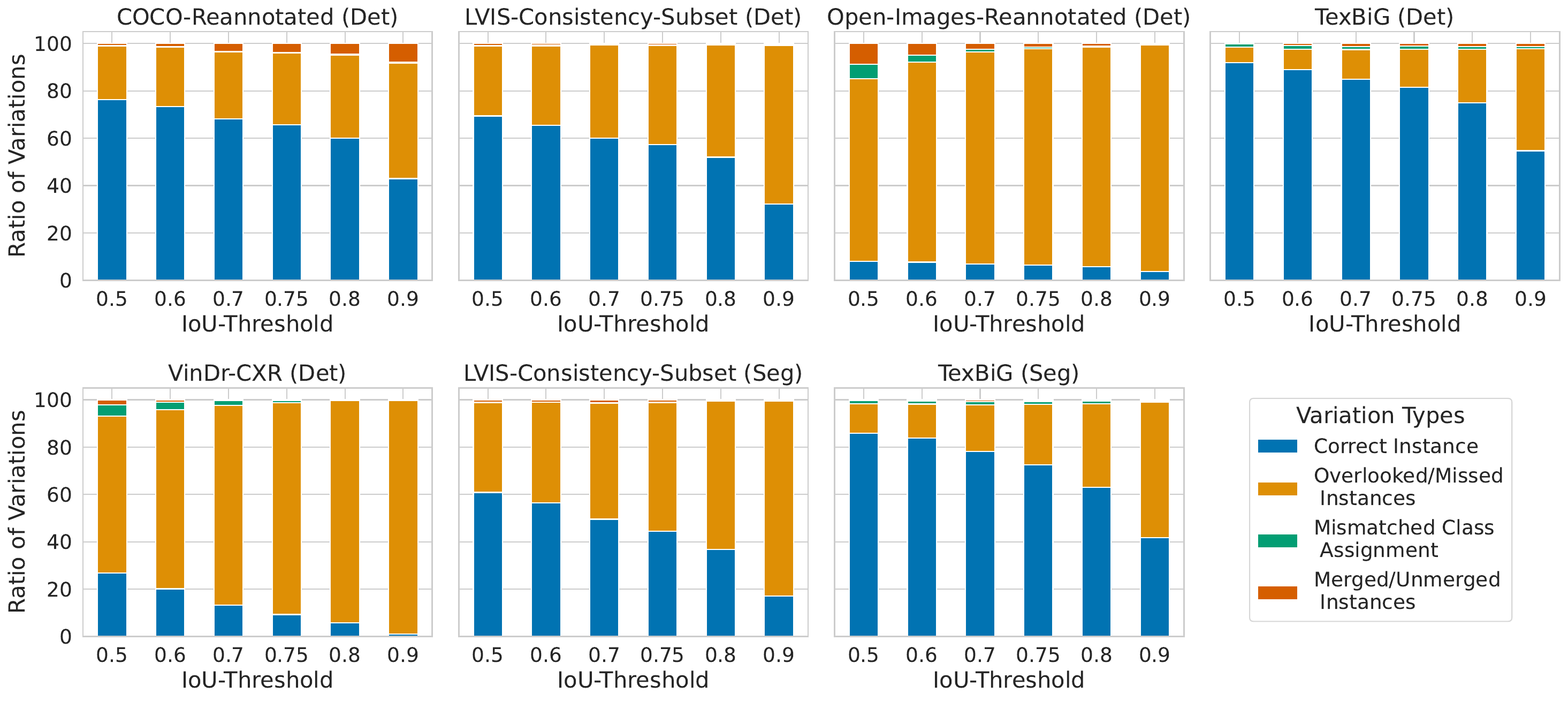}
    \caption{Variation distribution across the remaining datasets, with (Det) referring to object detection and (Seg) referring to instance segmentation, indicating similar trends to those observed with TexBiG and LVIS. However, these two datasets are of relatively high agreement with good annotation quality.}
    \label{fig:error-distribution-big}
\end{center}
\end{figure*}

\begin{enumerate}
    \item \textbf{Localization Overlap Calculation:} The intersection over union (IoU) is calculated between different annotators $r$ for each of their respective instances. For example, take annotator A and annotator B.
    \begin{equation}
        IoU(\Tilde{y}_{ij}^{A}, \Tilde{y}_{ij}^{B}) = \frac{\vert \Tilde{y}_{ij}^{A} \cap \Tilde{y}_{ij}^{B} \vert}{\vert \Tilde{y}_{ij}^{A} \cup \Tilde{y}_{ij}^{B} \vert}
    \end{equation}
    
    \item \textbf{Cost Matrix and Matching:} A cost matrix is created using the function:
    \begin{equation}
        C(j,k) = 1 - IoU(\Tilde{y}_{ij}^{A}, \Tilde{y}_{ik}^{B})
    \end{equation}
    Assume that annotator $A$ has $M_A$ annotations and annotator $B$ has $M_B$ annotations for image $i$. The sets are matched using the Hungarian algorithm, ensuring $M_A = M_B$ by padding the smaller set with $\varnothing$. For multiple annotators ($R > 2$), a greedy matching is algorithm is applied between the matched sets.
    
    \item \textbf{Reliability Data and Coincidence Matrix:} After matching, reliability data is organized into a coincidence matrix with values $o_{ck}$ representing the number of c-k pairs (referring here to a pair of categories assigned to the same unit by different annotators) for each instance (unit) $u$, calculated as:
    \begin{equation}
        o_{ck} = \sum_u \frac{\text{Number of c-k pairs in unit u}}{m_u-1}
    \end{equation}
    where $m_u$ is the number of annotators (observers) for unit $u$, so how many annotators found the same instance $u$. From this, we calculate:
    \begin{equation}
        n_c = \sum_k o_{ck} \quad \text{and} \quad n = \sum_c n_c
    \end{equation}
    Here, $n_c$ represents the total number of times category $c$ was assigned across all units, and $n$ is the total number of paired observations across all categories.

    \item \textbf{Krippendorff's Alpha Calculation:} Finally, $\alpha$ for nominal data is calculated using:
    \begin{equation}
        \alpha = 1- \frac{D_o}{D_e} = \frac{(n-1)\sum_c o_{cc}-\sum_c n_c (n_c -1 )}{n(n-1)-\sum_c n_c (n_c-1)}
    \end{equation}
\end{enumerate}

Further information about the method can be found in the paper \cite{tschirschwitz2022}.

\subsection*{Interpretation of Alpha Values}
\begin{itemize}
    \item $\alpha \geq 0.8$ signifies reliable and strong agreement among raters.
    \item $\alpha \geq 0.667$ is considered acceptable with moderate agreement.
    \item $\alpha = 0$ indicates agreement no better than chance, suggesting random assignment of classes.
    \item $\alpha < 0$ denotes systematic disagreement, which could indicate unclear guidelines, insufficient rater expertise, or particularly challenging images.
\end{itemize}

To ensure the accuracy of this method, the method discourages missing entries by replacing them with a filler class, leading to worse agreement scores if an annotator misses an entry that others found.

\section*{Appendix C: Additional Material - Annotation Variation Type Analysis}
\label{sec:appendix-C}

For counting the variations, we employ an algorithm designed to match as many instances as possible. The algorithm requires three elements for each image: 1) the annotations, 2) an IoU threshold, and 3) a list of annotators assigned to this image. For each possible pair of annotators, their respective instance IoU is calculated. Using this localization information:
\begin{enumerate}
    \item \textbf{Matching of Correct Instances:} Instances of the same class are matched starting with the highest overlapping pair of instances until the last pair with an IoU value greater than or equal to the IoU threshold. These instances are then excluded from further matching.
    \item \textbf{Matching of Merged/Unmerged Instances with Correct Classes:} In the next step, all remaining instances from each annotator are merged within their own class. These merged instances are then included in the IoU evaluation, and the same matching procedure is executed again, excluding possible matches.
    \item \textbf{Matching of Wrong-Class Instances:} Instances with correct localization but mismatching classes are matched next, following the same procedure, this excludes the previously merged instances.
    \item \textbf{Matching of Merged/Unmerged Instances with Incorrect Classes:} Similar to step 2, merged instances are created within the annotations of a single annotator but are now allowed to match with instances from other classes from the other annotator.
    \item \textbf{Missing/Additional Instances:} All remaining instances are counted as missing or additional, as they did not find any match.
    \end{enumerate}
With this hierarchical procedure, we aim to match as many instances as possible, essentially adopting a lenient approach toward annotation mistakes. This means that while an instance with a better overlap might be available, the chosen match will correspond to the class of the annotation. This approach maximizes agreement wherever possible. Therefore, matches with higher IoU are generally preferred, but matches with fitting classes take precedence if they exceed the IoU threshold.
\par
\begin{figure}[h]
    \centering
    \includegraphics[width=\linewidth]{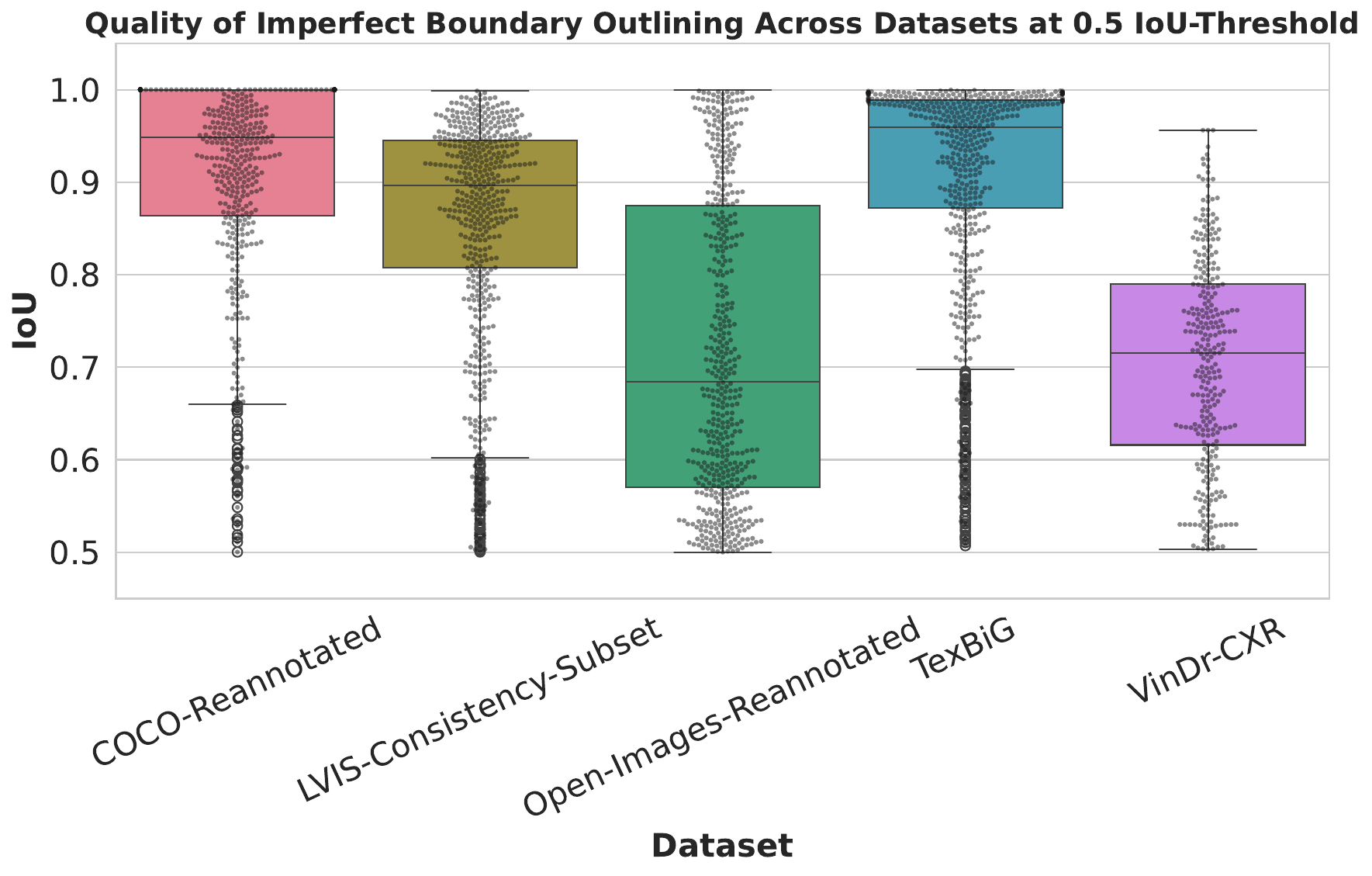}
    \caption{Boundary quality, illustrating how good the localization quality within correct classes is. COCO-Reannotated shows a very high number of 1.0 IoU overlaps, suggesting possible duplication from the original dataset to the reannotated version.}
    \label{fig:boundary-outline}
\end{figure}

In Figure \ref{fig:error-distribution-big}, we present the variation distribution across the different analyzed datasets. Figure \ref{fig:boundary-outline} visualizes the boundary qualities observed with an IoU threshold of 0.5. The remaining three images illustrate various types of variations.

\begin{figure}[ht]
\begin{center}
    \includegraphics[width=0.80\linewidth]{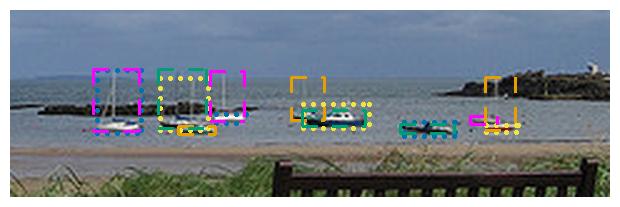}
    \caption{The dotted line represents annotator A while the dashed line represents annotator B. We can see that the boats are very hard to recognize when not zoomed into the image (full image \ref{fig:boats-full-image}. We consider this an annotation variation caused by image quality or at least perceived image quality, as this might also be related to the available tooling for the annotation process.}
    \label{fig:boats-zoomed}
\end{center}
\end{figure}

\begin{figure}[ht]
\begin{center}
    \includegraphics[width=0.80\linewidth]{figures/000000522713.jpg}
    \caption{This image shows a full image without any annotation, and Figure \ref{fig:boats-zoomed} a zoomed in version.}
    \label{fig:boats-full-image}
\end{center}
\end{figure}

\begin{figure}[ht]
\begin{center}
    \includegraphics[width=\linewidth]{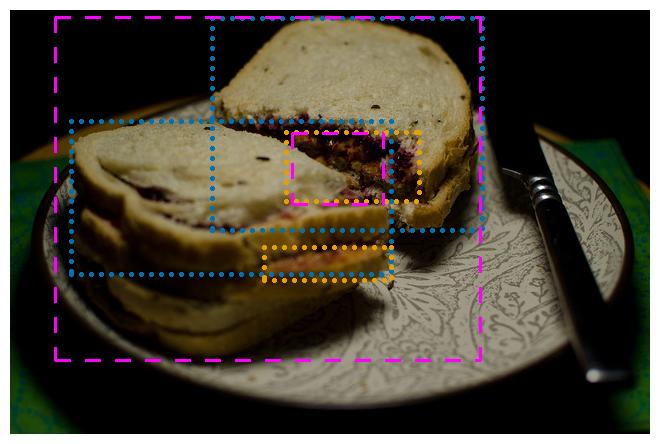}
    \caption{The dotted line represents annotator A while the dashed line represents annotator B. This image again shows a case of a merging issue, where both annotators made reasonable assumptions about the labeling convention, however the guideline seems to be not specific enough. The magenta instance in the center and the two orange instances are parts of the class peanut butter. Here the interpretation seems very difficult, almost like an occlusion case. One annotator opted for additional peanut butter at the bottom sandwich, while the other annotator did not find any peanut butter there. We also attribute this issue to image quality.}
    \label{fig:sandwiches}
\end{center}
\end{figure}

\end{document}